\documentclass{l4dc2024}
\usepackage[margin=1in]{geometry}
\usepackage[utf8]{inputenc}
\usepackage{graphicx, booktabs}
\usepackage[format=plain]{caption}
\usepackage{subcaption}
\usepackage{graphicx,wrapfig}

\usepackage{amsfonts}       
\usepackage{nicefrac}       
\usepackage{microtype}      
\usepackage{xcolor}   
\usepackage{amsmath}
\usepackage{amssymb}
\usepackage{mathtools}

\usepackage[ruled]{algorithm2e}

\newcommand{\plusminus}[1] {\scriptsize{$\pm$ #1}}

\usepackage{amsmath,amsfonts,bm}

\usepackage{xcolor}

\definecolor{bbblue}{rgb}{0.47843137254901963, 0.8392156862745098, 0.803921568627451}

\definecolor{mygreen}{RGB}{140,177,109}
\definecolor{myyellow}{RGB}{206,158,53}
\definecolor{mypurple}{RGB}{145,116,163}
\definecolor{myorange}{RGB}{233,113,45}
\definecolor{myblue}{RGB}{125,146,182}

\def\eqref#1{equation~\ref{#1}}

\def\1{\bm{1}}

\def\rvx{{\mathbf{x}}}
\def\rvy{{\mathbf{y}}}

\DeclareMathAlphabet{\mathsfit}{\encodingdefault}{\sfdefault}{m}{sl}
\SetMathAlphabet{\mathsfit}{bold}{\encodingdefault}{\sfdefault}{bx}{n}

\newcommand{\R}{\mathbb{R}}

\title[Domain Adversarial Neural Process]{Data-Driven Simulator for Mechanical Circulatory Support with \\Domain Adversarial Neural Process}

\usepackage{times}

\author{%
 \Name{Sophia Sun} \Email{shs066@ucsd.edu}\\
 \addr University of California, San Diego
 \AND
 \Name{Wenyuan Chen} \Email{wenyuanc@stanford.edu}\\
 \addr Stanford University 
 \AND
 \Name{Zihao Zhou} \Email{ziz244@ucsd.edu}\\
 \addr University of California, San Diego
 \AND
  \Name{Sonia Fereidooni} \Email{sfereidooni@ucsd.edu}\\
 \addr University of California, San Diego
 \AND
  \Name{Elise Jortberg} \Email{ejortberg@abiomed.com}\\
 \addr AbioMed
 \AND
  \Name{Rose Yu} \Email{roseyu@ucsd.edu}\\
 \addr University of California, San Diego
}

\begin{document}

\maketitle

\begin{abstract}%

We propose a data-driven simulator for Mechanical Circulatory Support (MCS) devices, implemented as a probabilistic deep sequence model. 
Existing mechanical simulators for MCS rely on oversimplifying assumptions and are insensitive to patient-specific behavior, limiting their applicability to real-world treatment scenarios. To address these shortcomings, our model Domain Adversarial Neural Process (DANP) employs a neural process architecture, allowing it to capture the probabilistic relationship between MCS pump levels and aortic pressure measurements with uncertainty. We use domain adversarial training to combine simulation data with real-world observations, resulting in a more realistic and diverse representation of potential outcomes. Empirical results with an improvement of 19\% in non-stationary trend prediction establish DANP as an effective tool for clinicians to understand and make informed decisions regarding MCS patient treatment. The code for DANP is open-sourced at \url{https://github.com/Rose-STL-Lab/DANP}.









\end{abstract}


\section{Introduction}



Machine learning has shown great promise to revolutionize patient care by improving outcome prediction and supporting clinical decisions \citep{davenport2019potential}.  However, it still faces significant challenges in learning accurate dynamics models from medical time series data, due to their non-stationary, uncertain, and confounding nature \citep{gottesman2019guidelines}.  There is also a distribution shift issue when sequentially suggesting optimal treatment plans, as the predicted future scenarios may not be in the training data.

In this paper, we aim to simulate the impact of Mechanical Circulatory Support (MCS) support levels on patient blood pressure. Forward flow MCS devices are mechanical devices designed to assist the heart in pumping blood from the left ventricle into the ascending aorta to deliver oxygenated blood to the body. They ensure that blood continues to circulate properly throughout the body when the heart fails to perform this function adequately. MCS devices play a pivotal role in maintaining blood pressure, maintaining organ perfusion and aiding heart muscle recovery.

Accurate simulation of MCS support is highly valuable for clinical decision support in treatment plan optimization. The simulator would allow physicians to understand how different MCS support levels would  change blood pressure in a hypothetical setting.  Traditionally, while clinicians can control the support level of MCS, there are limited recommendations and no universal protocols for optimally operating or weaning the device \citep{escalation}. In practice, clinicians often rely on empirical experiences, which are difficult to transfer or standardize. A simulator would allow clinicians to ask ``what-if'' questions, providing critical information of plausible scenarios to consider before implementing the treatment change.

While there exist mechanical simulators for MCS support levels \citep{schampaert2011vitro}, they are based on over-simplified assumptions of the human arterial system \citep{bonfanti2017computational}.
Since the simulation only relies on mechanistic parameters as input, they are insensitive to patient-specific behavior \citep{van2019comparing}, making them less applicable in real-life treatment practices. Furthermore, biomechanical simulators are deterministic and cannot capture the uncertainty common in the treatment of MCS patients.  Finally, many of the simulation parameters are theoretical and not readily observable in regular practice. Directly training a simulation model on real-world data would suffer from the distribution shift issue as many of the risky treatments would not appear in the training set.

Our paper presents a data-driven simulator for Mechanical Circulatory Support (MCS) using data from the  \textit{Impella CP} device \citep{impella2014instructions}. 
We developed a novel model, Domain Adversarial Neural Process (DANP), to address these challenges. DANP employs an encoder-decoder neural process model to learn the probabilistic relationship between the MCS pump support level (P-level) and key aortic pressure measurements. The recurrent neural network architecture of the encoder allows it to take varying context lengths as input. It can be conditioned on future P-level, allowing the user to ask ``what if'' questions. The neural process architecture can capture the uncertainty in the forecast. To mitigate the gap between simulation data and real world observations, we use domain adversarial training \citep{ganin2016domain} to incorporate diverse simulation data and enable sim-to-real transfer.

Empirical results from DANP validate its efficacy, showcasing substantial improvements in forecasting accuracy and trend prediction compared to existing methods. The outcome forecasts generated by DANP prove to be more realistic and diverse, addressing the limitations of previous simulators. Our model acts effectively as a data-driven simulator and provides clinicians with an effective tool with confidence boundaries for optimal MCS treatment planning. 

In summary, our contributions are:
\begin{itemize}
\itemsep0em 
\item We propose a data-driven simulator for MCS devices, implemented as a probabilistic deep sequence model, domain-adversarial neural process (DANP).

\item We present a novel sim-to-real transfer approach to integrate both simulated and real-world data from MCS devices. This allows our DANP model to answer “what-if” questions and generate diverse scenarios that are rarely seen in the data. The probabilistic nature of our model also helps clinicians to better address uncertainty in the treatment of MCS patients.

\item We provide qualitative and quantitative experiment results to show the efficacy of our model. MAP forecasts by DANP are $5\%$ more accurate overall and $19\%$ more accurate for non-stationary data compared to the best-performing baseline. DANP's more expressive and accurate forecasts can provide more insights compared to existing models and simulators.
\end{itemize}

\section{Related Works}
\paragraph{Existing simulators used in MCS.}
Traditional biomechanical simulators are based on over-simplified assumptions of the human arterial system \citep{bonfanti2017computational}. A significant gap exists between what happens in a simulator and what works in the real world. 
Furthermore, biomechanical simulators are deterministic. They cannot capture the uncertainty common in the treatment of MCS patients. Directly training a simulation model on real-world data would suffer from the distribution shift issue, as many of the risky treatments would not appear in the training set. 

Designing a hypothesis-driven healthcare simulator that encompasses all possible information is challenging \citep{gottesman2019guidelines}, and even leading synthetic data generation methods may miss some deviations, leading to invalid data \citep{chen2019validity}. 
For instance, \citet{prakosa2012generation} uses the traditional motion simulator to generate visually realistic cardiac images, while \citet{mahmood2018deep} improves the simulator trained on synthetic data using rendered real-world data. However, less attention has been paid to improving the traditional cardiac motion model's ability to capture accurately.
While intricate simulators have been developed for hemodynamic modeling of patient phenotypes and cardiovascular treatment \citep{harvi}, they rely on strong modeling assumptions and require clinical validation, necessitating the development of improved approaches that bridge the gap between mechanistic (hypothesis-driven) and phenomenological (real-world data-driven) models within the healthcare domain
\citep{frangi2018simulation}.

\paragraph{Time series forecasting in the medical applications.} Most works that address medical time-series forecasting focus their efforts on large-scale electronic health records (EHRs);
we refer the readers to \citep{morid2023time} for a detailed survey. Many works use Gaussian processes to impute and model multivariate medical time series data \citep{schulam2015framework, cheng2017sparse, prasad2017reinforcement} because they naturally capture irregular time series observations and can estimate prediction uncertainties in a probabilistic framework \citep{roberts2013gaussian}. In this work, we use a Neural Process model \citep{garnelo2018neural} that combines Gaussian processes with deep neural networks. It inherits many desirable properties of GP but is more expressive and efficient to optimize. 


\paragraph{Sim-to-real adaptation.} Developing data-driven simulators has been mostly prominent in robotic applications \citep{zhao2020sim}. Recent works have shown that by transferring knowledge learned in simulation towards their deployment in the real world, domain adaptation can improve the model's convergence and robustness for a diverse set of tasks \citep{cheng2019end}. Common approaches include meta learning \citep{arndt2020meta}, continual learning \citep{traore2019continual}, and using domain-adversarial training \citep{ganin2016domain} to learn domain-agnostic feature representations. The studies of sim-to-real transfer to build better simulators for medical applications remain scarce.

\paragraph{Background of Impella.} The Impella CP is a left ventricular assist device (LVAD) that supports the heart’s function via a percutaneously inserted catheter. It pumps blood from the left ventricle (LV) to the ascending aorta. The device has positive effects on cardiovascular parameters, including cardiac output, coronary blood flow, LV end-diastolic pressure, and myocardial workload. The Automated Impella Controller (AIC) provides continuous high-resolution data at 25 Hz to monitor aortic pressure, motor current, motor speed, and pump flow. One of the primary indications for the use of Impella is during high-risk percutaneous coronary interventions (HR-PCI) \citep{impella2014instructions}. A PCI is a cardiac catheterization lab-based procedure to address coronary artery disease (CAD). PCIs considered high risk are a function of cardiovascular risk factors, including the complexity of CAD, hemodynamic compromise, and patient comorbidities. The Impella CP is used in HR-PCI for better clinical outcomes and to avert hemodynamic instability.

\section{Problem Setting and Data}

\begin{figure}
    \centering
\includegraphics[width=0.52\linewidth]{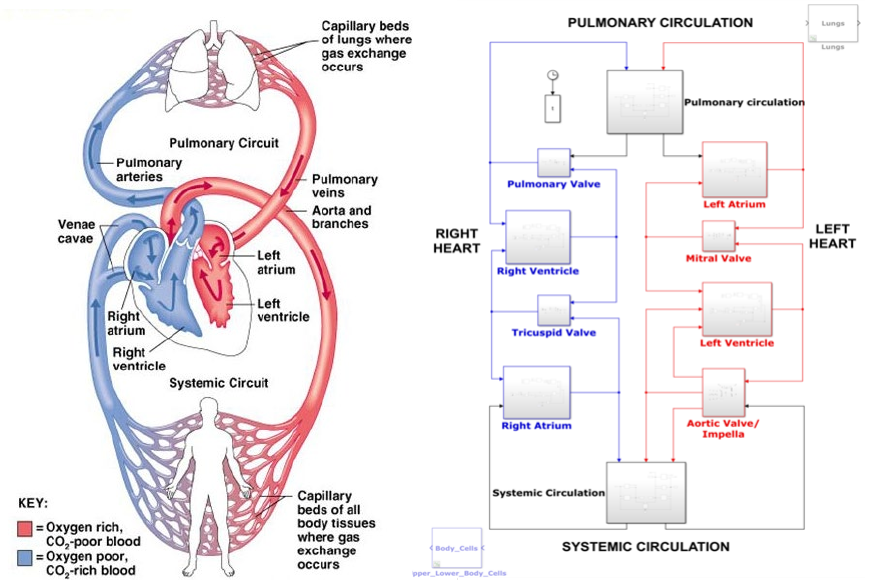}
\label{fig:eq0}\includegraphics[width=.47\textwidth]{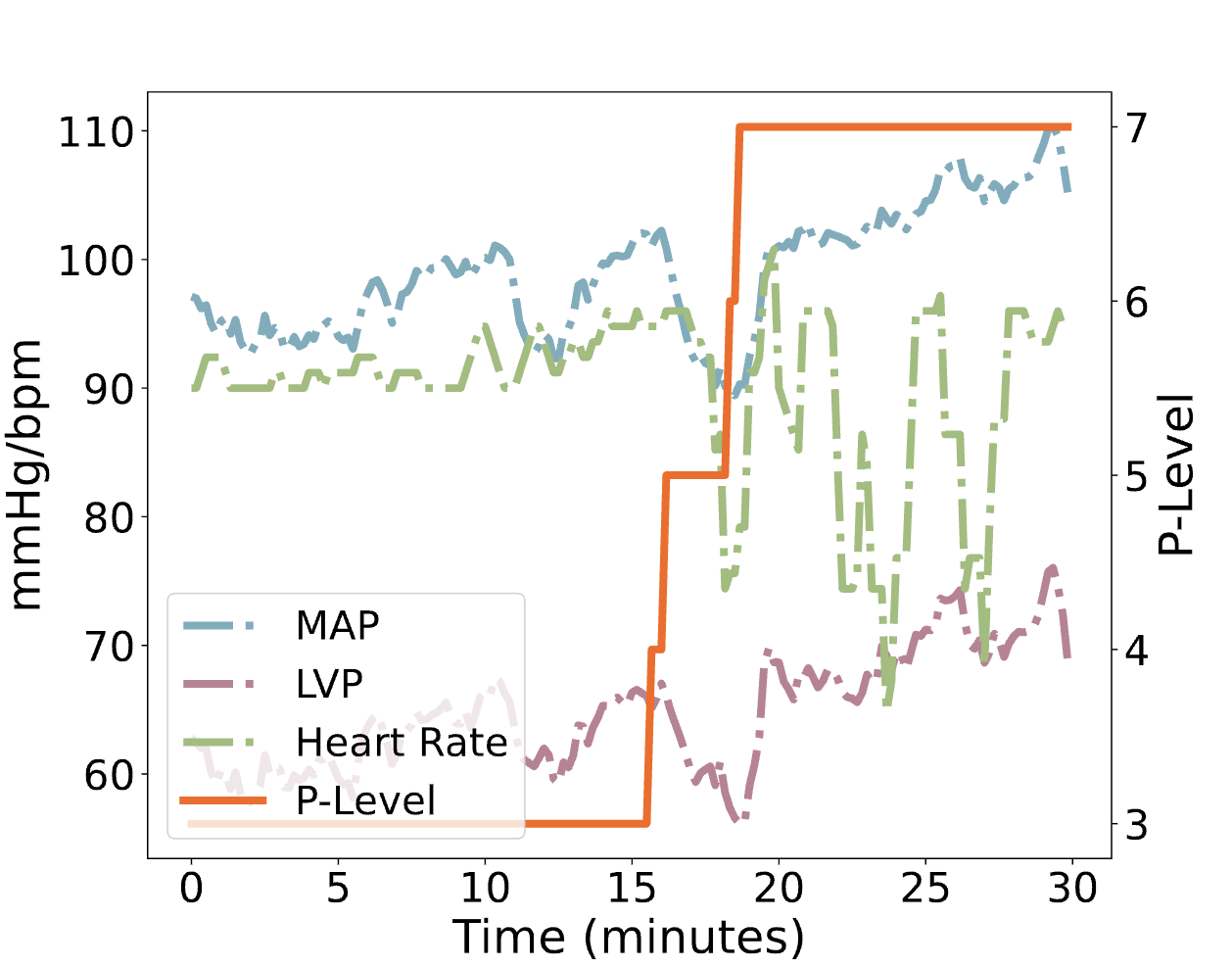}
\caption{ \textbf{Left}: Cardiovascular System overview (adapted from Pearson's ``Essentials of Human Anatomy and Physiology'' Ch. 11 ``The Cardiovascular System''. \textbf{Middle}: Translation to Lumped Parameter circuitry model, the source of our simulated data. \textbf{Right}: Visualizations of a patient data sample. Note that as the clinician increases the \textcolor{myorange}{pump level}, the \textcolor{myblue}{mean aortic pressure} (MAP) increases as a result.}
\label{fig:LP}
\end{figure}

Our work models the effect of different levels of MCS support on patients' hemodynamic status. We provide a visualization of the system of interest in Figure \ref{fig:LP} (left). The main variable to be predicted is future Mean Aortic Pressure (MAP), a key hemodynamic parameter that serves as a critical endpoint for assessing organ perfusion and determining the readiness for de-escalation of support. 
The desiderata of our data-driven simulator are: (1) to accurately forecast future MAP based on past data, and (2) to be able to reason about alternative input of pump speed and produce sensible hypothetical outcomes.

Mathematically, the simulator is a function that maps the historical MAP along with other hemodynamic features and the proposed future P-levels to future MAP values. We can define the problem as a multi-step time series forecasting task. Let $\rvx = x^{1:k} = (x^1, \ldots, x^k)  \in \R^{k \times d_x}$ 
denote the $k$ time steps of $d_x$ input features, $\textbf{pl} = pl^{1:h} = (pl^1, \ldots, pl^h) \in \{1, \ldots, 9\}^{h}$ the hypothetical pump support level (P-level) of the future $h$ steps, and $\rvy = y^{1:h} = (y^1, \ldots, y^h) \in \R^{h}$ as the  MAP values of the future $h$ time steps. The superscript indicates the data time step. Our model learns the function:
\begin{equation}
  f:  x^{1:k} , pl^{1:h} \longrightarrow y^{1:h} 
\label{eqn: problem}
\end{equation}

The model takes an input sequence of 15 minutes ($k=90$) and outputs a sequence of the next 15 minutes ($h= 90$). The horizon lengths and are chosen based on suggestions from physicians. For training and testing our models, the ``future'' support level aligns with the actions taken by physicians in real-life scenarios, enabling us to calculate the errors in MAP prediction retrospectively. We selected the following $d_x = 7$ features as simulator input, based on medical relevance and their availability for potential real-time decision-making. Figure \ref{fig:LP} (right) visualizes some features of a real data sample.
\begin{enumerate}
\itemsep0.1em 
    \item \textbf{Aortic Pressure (MAP)}: The Aortic Pressure (mmHg) is the blood pressure at the aorta's root and is directly measured by an optical sensor on the Impella CP's outlet shown in Figure \ref{RT25} in the Appendix. 
    \item \textbf{P-level (Motor Speed)}:  Impella operates at 9 different speed levels, from P1-P9, each with a constant motor speed (rpm). The P-level proportionally determines the blood flow provided to the patient by the motor's speed and current. Clinicians can control the P-level
    while the patient is on support.
    \item \textbf{Pump Flow}: (L/min) is the calculated forward blood flow through the Impella cannula.
    \item \textbf{Left Ventricular Pressure (LVP)}:  (mmHg), the pressure inside the heart's left ventricle during each cardiac cycle.  \citep{understanding}. 
    \item \textbf{Heart Rate}: (BPM) Heart rate is estimated from Aortic Pressure over the cardiac cycle via a smooth filter and Fourier transform.
     \item \textbf{Tau\_LV}: (ms) Time constant of isovolumic relaxation ($\tau$) representing the time LVP takes to reduce by 1/e from aortic valve closure. Estimated from an exponential fit on the downstroke of the LVP curve.
     \item \textbf{Contractility}: A measure of load-dependent ventricular contractility (mmHg/s), estimated using a linear transformation of the maximum rate of LV pressure rise.
\end{enumerate}

\subsection{Simulated Data}

We utilize data generated from an electrical circuit-inspired simulator. The Windkessel  Lumped Parameter circuitry simulator \citep{windkessel}  models cardiac function through a parameterized set of differential equations. In the electrical circuit analogy, potentials represent the differential blood pressure, and currents represent blood flow, see the chart in the center of Figure \ref{fig:LP}. 

Among the input parameters in the cardiac simulator, four were determined to be derivable from sensor data within an Impella CP: P-level, heart rate, contractility, and the time constant of relaxation (Tau). A grid search was performed across the physiological ranges for these four parameters and passed into the simulator. 
The simulator outputs included MAP, pump flow rate, and the Left Ventricular Pressure (LVP).  An artificial time series was created from the simulator point estimates by sampling from the simulator with random noise. Data augmentation techniques of shifting and rescaling were performed on the time-series samples to imitate the effect of unused input parameters from the simulator.


\subsection{Real world data}

The simulator alone does not adequately address clinical decision support during patient treatment, as many of the input parameters are theoretical and not readily observable in regular practice. In training our data-driven simulator, we perform a sim-to-real transfer using both simulated data and real data.

Real-world data was sourced from 211 High-Risk Percutaneous Coronary Intervention (HR-PCI) patients. Among the features, heart rate, end-systolic elastance, and the time constant of relaxation $\tau$ were derived features from the 25Hz signals. The raw waveform data from the Impella pump are sampled at 25 Hz, and we process the raw waveform data into a rolling average of 0.1 Hz time series.  We filter out samples with outlier features and those with medication or intervention alarms to avoid confounding. In total, the HR-PCI data set consists of 43,814 30-minute-long time series.



\section{Domain Adversarial Neural Process (DANP)}

We present Domain Adversarial Neural Process (DANP), a probabilistic deep sequence model for MCS simulation. We tailor the approach with a custom-designed Neural Process encoder and a sequence decoder for time series forecasting. Our method effectively bridges the domain differences between simulation and real-world data, acting as an adept data-driven simulator. 

\paragraph{Domain-Adversarial Training.} 
The high-level idea behind domain-adversarial training \cite{ganin2016domain} is to learn domain-invariant features by simultaneously optimizing two tasks: the primary prediction task (e.g., classification/regression) and a domain classification task. This is achieved through a shared feature extractor followed by a Gradient Reversal Layer (GRL) that encourages the network to learn features invariant to domain differences. In the context of our application, the simulation data helps the model learn from a diverse set of scenarios not present in the real dataset, without overly dominating the training data. The objective function for domain adversarial training is as follows: 
\begin{equation}
    L_{\text{DANP}} = L_{y} + \lambda \cdot L_{d}
    \label{eqn:DANP_loss}
\end{equation}
where \(L_{y}\) is the task loss in the target domain (i.e. the task of forecasting MAP), \(L_{d}\) is the domain classification loss (between simulation and real data), and \(\lambda\) is a hyperparameter controlling the trade-off between domain adaptation and task performance. 

\begin{figure*}[htbp]
\centering
\includegraphics[width=0.7\linewidth]{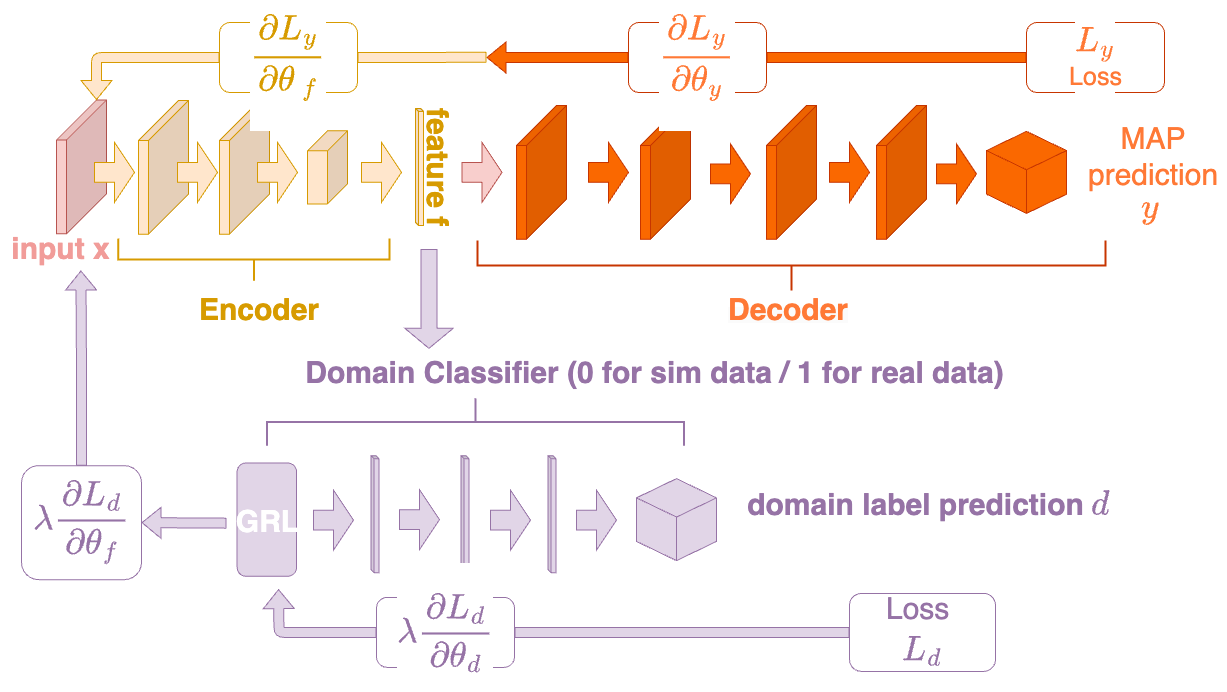}
\caption{ Architecture of DANP with the neural process \textcolor{myyellow}{encoder}, \textcolor{myorange}{decoder}, and a \textcolor{mypurple}{domain classifier}. Input \(\rvx\) (along with P-level $\mathbf{pl}$, combined for conciseness) is passed into the \textcolor{myyellow}{encoder}.  The output (feature \(z\)) is then passed separately into the \textcolor{mygreen}{decoder} and \textcolor{mypurple}{domain classifier} to obtain regression \({\rvy}\) prediction (MAP) and domain prediction \({d}\) (0 or 1), respectively. The Gradient Reversal Layer (GRL) for the domain classifier acts as an identity transform. The loss of \(y\) prediction (\(L_{y}\)) and domain prediction (\(L_{d}\)) will be calculated separately but backpropagated together.}
\label{DANP}
\end{figure*}

\paragraph{Neural Process.} Neural processes (NPs) \citep{garnelo2018neural} are a class of models that parameterize a stochastic process with neural networks. They are structured similarly to a Variational Autoencoder (VAE) \citep{kingma2013auto}  where a latent variable model is employed to estimate the conditional distribution over the outputs of the unlabeled (target) points given the set of labeled (context) points. We show the model architecture of DANP in Figure \ref{DANP}, for which we will discuss in detail in the following sections.


\begin{minipage}{0.65\textwidth}
  \centering
{\includegraphics[width=\textwidth]{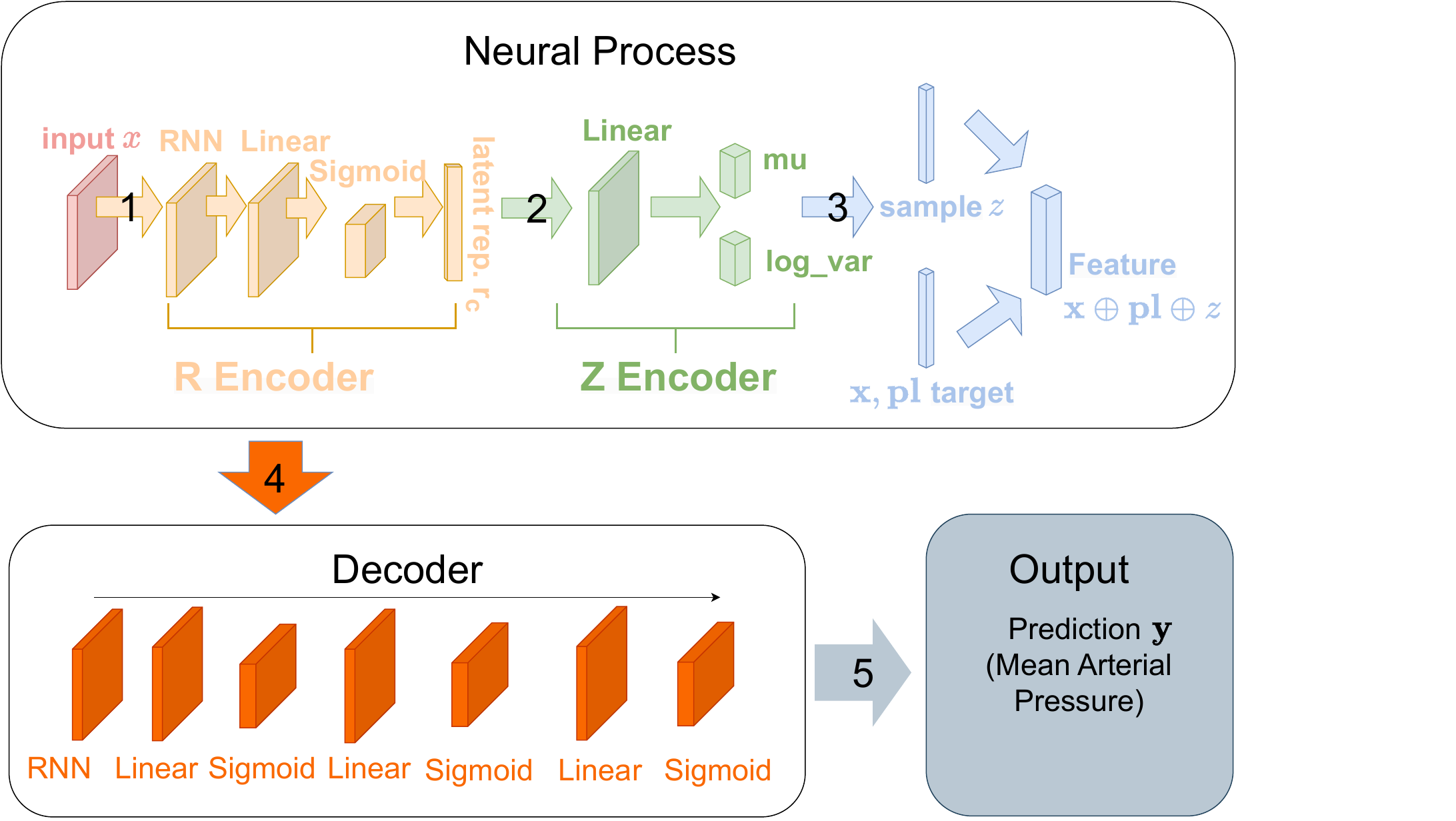}}
\end{minipage} 
\begin{minipage}{0.30\textwidth}
\centering
\includegraphics[width=\textwidth]{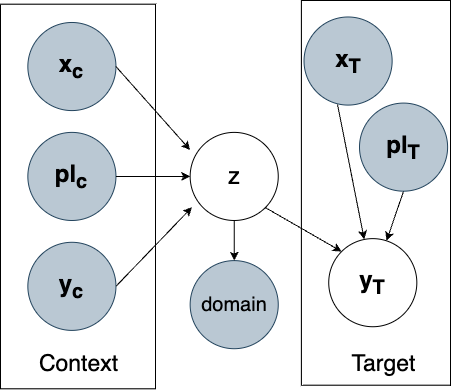}
\end{minipage}
\captionof{figure}{\textbf{Left: } Neural Process Architecture Details and \textcolor{myorange}{Decoder} Details. The input data \(\boldsymbol x\) is initially processed by an \textcolor{myyellow}{R Encoder} to obtain its latent representation. This latent representation is then further processed by a \textcolor{mygreen}{Z Encoder}, which parametrizes it into the \textcolor{myblue}{latent space \(z\)}. When concatenated with the target \(\boldsymbol x\), the sampled values from this latent space result in a combined feature \(\rvx_T \oplus \mathbf{pl}_T \oplus z  \). This is passed through a sequential \textcolor{myorange}{Decoder}, where it undergoes a series of transformations to ultimately generate predictions for MAP. \textbf{Right: } Graphical model of DANP.  A grey background indicates that the variable is observed.}
\label{fig:NP}

\vspace{1em}

\paragraph{Neural Process Encoder.} DANP's neural process encoder converts the input sequence into latent representations as the \textcolor{myorange}{Decoder} in Figure \ref{DANP}. The neural process model captures the intrinsic stochasticity of the data in a distribution over latent variables. This unique composition permits the neural process to derive a predictive distribution over functions, reflecting the uncertainty in patient data.

Specifically, the neural process encoder model learns a variational posterior distribution  $q_{\phi}(z|\boldsymbol x_c, \mathbf{pl}_c, \boldsymbol y_c )$ over a latent variable \( z \), based on a set of context (denoted as $\mathcal{C}$): 
\( (\boldsymbol x_c, \mathbf{pl}_c, \boldsymbol y_c ) =  \{ (\boldsymbol x_i,\mathbf{pl}_i, \boldsymbol y_i ) \}_{i\in \mathcal{C}} \), and parameterised by a neural network $\phi$. In training, each batch of data is randomly split 0.9/0.1 into context and target data. We use subscript here to index samples, and ${i\in \mathcal{C}}$ indicates that the sample is used as a context point. At the testing stage, all training data are used as context samples, and the test data as target samples.

We adopt a two two-part encoder design consisting of a sequential model and a distribution embedding model, as illustrated detailedly in Figure \ref{fig:NP}. 
The \textcolor{myyellow}{R-Encoder}  is a recurrent neural network (RNN) sequential model that transforms the sequence data into a representation vector \( r \). Other than dimension reduction, the recurrent architecture has the advantage of enabling us to potentially encode longer or shorter history for prediction.
The \textcolor{myyellow}{R-Encoder} takes in the context data $ (\boldsymbol x_c, \mathbf{pl}_c, \boldsymbol y_c )$, target point inputs  $ (\boldsymbol x_T,\mathbf{pl}_T) $, and yields the latent function representation $r$ as depicted in Figure \ref{fig:NP}. 

This latent representation \( r \) then serves as input to the \textcolor{mygreen}{Z-Encoder} implemented as a multi-layer perceptron. We assume that $z$ follows a multivariate standard normal distribution that captures the uncertainty in the function space. \textcolor{mygreen}{Z-Encoder} then outputs the mean $\mu_{z}$ and variance $\sigma^2_{z}$ of the variational distribution:
\[ q(z|\boldsymbol x_c, \mathbf{pl}_c, \boldsymbol y_c) = \mathcal{N}(\mu_{z}(r), \sigma^2_{z}(r)) \]


The concatenated feature of \( \rvx_T \oplus \mathbf{pl}_T \oplus z \) plays a dual role: it is used by the Domain Classifier to distinguish between simulation and real data, and also by the decoder to make predictions on $\rvy$. The domain classifier utilizes a gradient reversal function (GRL) on feature \(\rvx_T  \oplus \mathbf{pl}_T \oplus z\) before directing the output to a domain classifier, as shown in Figure \ref{DANP}. 

\paragraph{Sequential Decoder.} 

We captured the complexity of the prediction  with the \textcolor{myorange}{Sequential Decoder} in Figure \ref{DANP} (also shown in more detail as the \textcolor{myorange}{Decoder} of Figure \ref{fig:NP}), a neural network model $\theta$ that parameterized 
\[ p_{\boldsymbol \theta}(y_{n+1}| \rvx_T, \mathbf{pl}_T, z) = \mathcal{N}(\mu_\theta (\rvx_T  \oplus \mathbf{pl}_T \oplus z), \sigma^2_\theta (\rvx_T  \oplus \mathbf{pl}_T \oplus z))\]
for a new observation \( y_{n+1} \). 
The \textcolor{myorange}{Sequential Decoder} is implemented as two RNN layers, followed by four linear layers with a sigmoid activation function. 

In the training phase, the neural process model maximizes the probability over the evidence lower bound (ELBO) loss, defined as 
\begin{equation}
\text{ELBO}(\theta, q) = \mathbb{E}_{z \sim q(z;\rvx_c, \mathbf{pl}_c, \rvy_c)} [\log p(\rvx_T, \mathbf{pl}_T|z;\theta)] \nonumber - \text{KL}[q(z;\rvx_c, \mathbf{pl}_c, \rvy_c) || p(z|\rvx_T, \mathbf{pl}_T;\theta)]
\label{eqn:elbo}
\end{equation} 

where \( \theta \) are the model parameters, \( z \) indicates latent variables, and \( \rvx, \mathbf{pl}, \rvy\) represents observed data in the context and target sets. This is the $L_y$ loss in equation \ref{eqn:DANP_loss}. During inference, a vector $z$ is sampled from encoded $p(z)$. The architecture resonates with the depiction in Figure \ref{fig:NP}, wherein the x target within the neural process amalgamates features of the input sequence and the future sequence's feature - the P-level, to generate the feature \(\rvx_T  \oplus \mathbf{pl}_T \oplus z\). This is then channeled to the decoder to predict \( \rvy_T \).

\paragraph{Domain Classifier.} As demonstrated in the purple part in Figure \ref{DANP}, the real data is labeled as 1, and the sim data is labeled as 0. Similar to the sequential decoder, the feature \(\rvx_T  \oplus \mathbf{pl}_T \oplus z\), extracted by the Neural Process, undergoes a series of transformations. It begins with a gradient reversal layer(GRL) that multiplies the gradient by a certain negative constant during back-propagation. It effectively makes the network learn features that are insensitive to domain-specific variations. The output from GRL is passed to a series of transformations in the domain classifier, which is responsible for classifying the domain of the input data. The domain classifier learns to distinguish between features from different domains. 

To train the model, we employ the loss function in \eqref{eqn:DANP_loss}. First, the ELBO loss, represented as \(L_{y}\), is employed to reduce the gap between the model's predictions and the actual MAP values. It is the sum of the reconstruction loss, Mean Absolute Error (MAE), and the regularization term, the KL divergence between the approximate posterior and the prior of $\bold z$. 
The domain classification loss \(L_{d}\) is a Negative Log Likelihood Loss (NLLLoss) applied to the predicted domain labels. We aggregate these losses and back-propagate them jointly, as illustrated in Figure \ref{DANP}. This training approach equips the sim-to-real model with the capability to generate outputs that closely mirror real-world scenarios, exhibiting minimal error with uncertainty.





\section{Experiments}

\subsection{Baselines}
We compare our model to the following baselines:
\vspace{-0.5em}
 \begin{itemize}
\itemsep0em 
     \item \texttt{MLP}: Multi-layer perceptron neural network that takes all the input time series data and outputs the prediction both as a flattened vector.
     \item \texttt{CLMU:}  \cite{li2022forecasting}
    Conditional Legendre Memory Unit (LMU) that employs an LMU encoder \citep{voelker2019legendre} with a neural process decoder trained on both simulation and real data. The conditional LMU has a fixed initial hidden state based on the MAP standard deviations from the specific domain and encodes the initial time steps of a sample.  During training, real data was incorporated, and the label ``simulated" or ``real" was passed as a feature to the CLMU encoder as a condition.
     \item \texttt{NP direct transfer}: DANP's neural process model, trained only on \textit{simulation} data, without domain adaptation. 
    \item \texttt{NP no sim}: DANP's neural process model, trained only on \textit{real} data, without domain adaptation.
 \end{itemize}
 \vspace{-0.5em}
\subsection{Experiment Results}

\paragraph{Prediction Accuracy.}
We present the result of our experiments in Table \ref{tab:main_results}. To better evaluate the behavior and expressively of the models, we split the test data into three trend categories: increasing (inc), decreasing (dec), or stationary (stat), and report the MAE calculated on each category as well. In both of our training and testing sets, stationary streams make up for around 65\% of the data, increasing trends 20\%, and decreasing trends 15\%. We refer the readers to Appendix \ref{app:data} for details on how the trends are calculated.

Overall, DANP showed superior accuracy in MAP prediction  in all our metrics, improving $20\%$ over the best baselines for increase-trending data and $13\%$ in terms of trend prediction accuracy. Such improvements are significant because the non-stationary scenarios are clinically actionable and for which an accurate simulator is most helpful. We also show in Appendix \ref{app:exp} a data distribution analysis showing that DANP is able to better match the distribution of real data compared to the baseline models.

 \begin{table}[htbp]
     \centering
     \begin{tabular}{l c c c c c}
     \toprule
         \textbf{Method} &  \textbf{MAE (mmHg)} $\downarrow$ &   \textbf{MAE (inc)} $\downarrow$ &  \textbf{MAE (dec)} $\downarrow$ & \textbf{MAE (stat)} $\downarrow$ &  \textbf{Trend Acc} $\uparrow$ \\ 
    \midrule
      \texttt{MLP} & 7.97 \plusminus .26 & 9.04 \plusminus .68& 10.96 \plusminus .61&  6.78 \plusminus .43 & 0.57 \plusminus .03\\
       \texttt{CLMU} & 6.93 \plusminus .11  &  8.65 \plusminus .56  & 8.47 \plusminus .24 &  5.51 \plusminus .04 & 0.65 \plusminus .01 \\
    \texttt{NP\scriptsize{ direct transfer}} & 7.36 \plusminus .91 & 9.72 \plusminus 1.23 & 8.79 \plusminus 1.06 & 6.25 \plusminus .95 & 0.64 \plusminus .00 \\
      \texttt{NP\scriptsize{ no sim}} & 8.68 \plusminus .06 & \textbf{6.90} \plusminus .01 & 15.34 \plusminus .02 &  7.63 \plusminus .01 & 0.52 \plusminus .00\\
       \texttt{DANP} (ours) & \textbf{6.65} \plusminus .13 & \textbf{6.94} \plusminus .10 & \textbf{8.46} \plusminus .17 &  \textbf{ 5.36} \plusminus .09 & \textbf{0.70} \plusminus .01
 \\
    \bottomrule
     \end{tabular}
     \caption{Empirical results in terms of Mean Average Error (MAE) for data with increasing (inc), decreasing (dec), stationary (stat) trends, and trend prediction accuracy. DANP achieves significantly lower performs significantly better on trending data compared to baselines. }
     \label{tab:main_results}
 \end{table}

\begin{figure}[t!]
    \centering
    \subfigure[NP \scriptsize{direct transfer}]{\includegraphics[width=0.32\linewidth]{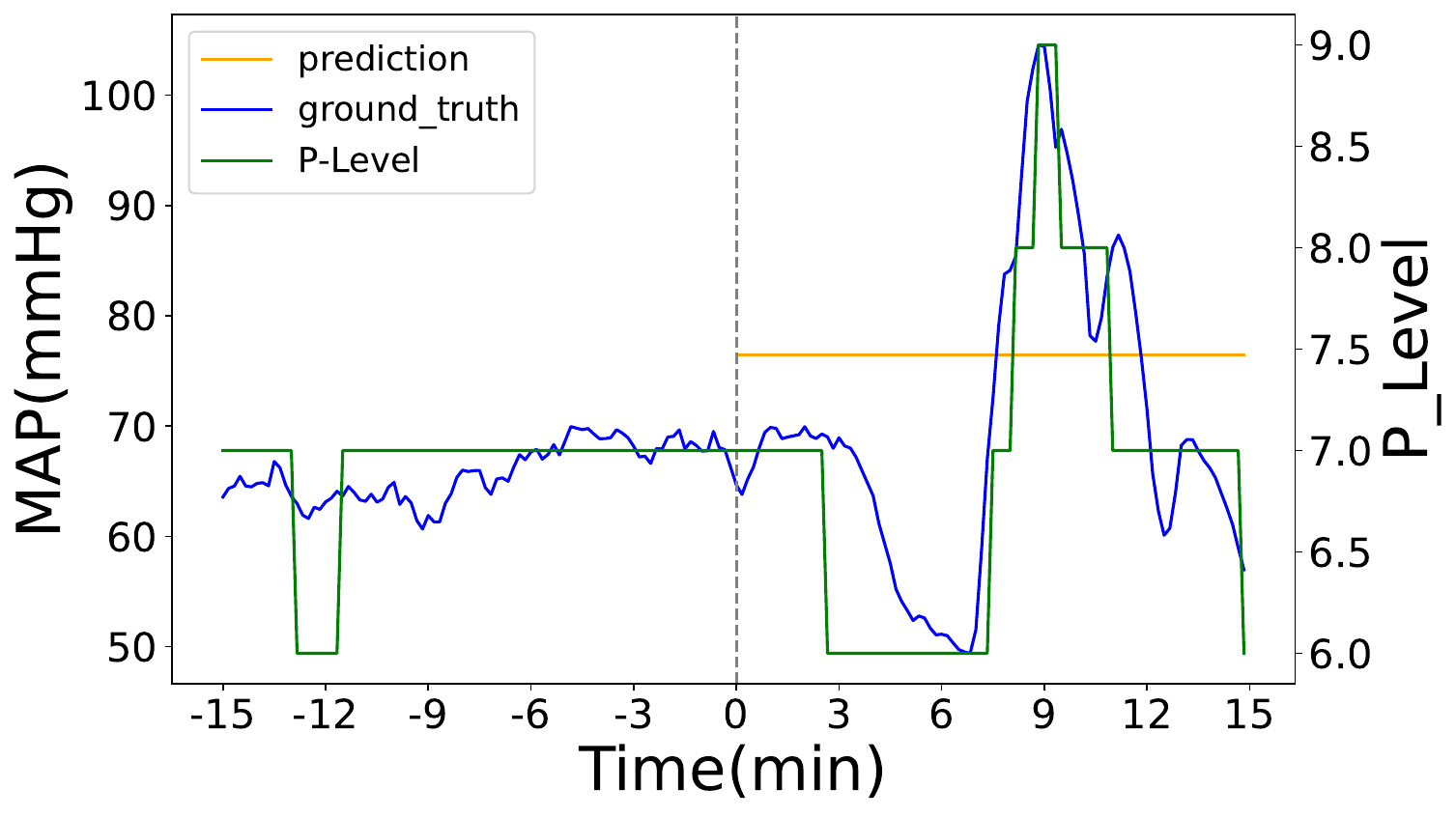}}
    \subfigure[CLMU]{\includegraphics[width=0.32\linewidth]{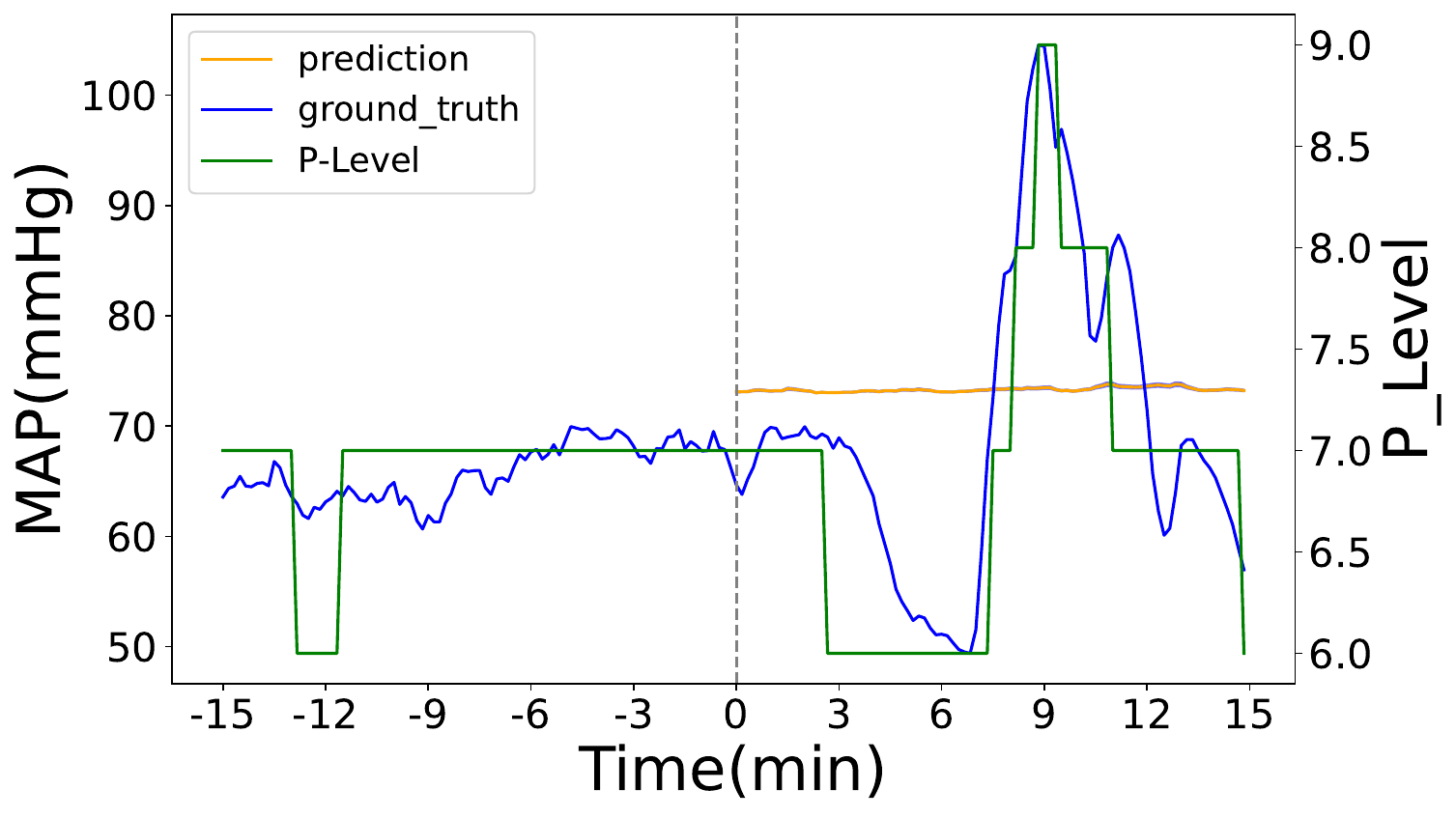}}
    \subfigure[DANP \textbf{(ours)}]{\includegraphics[width=0.32\linewidth]{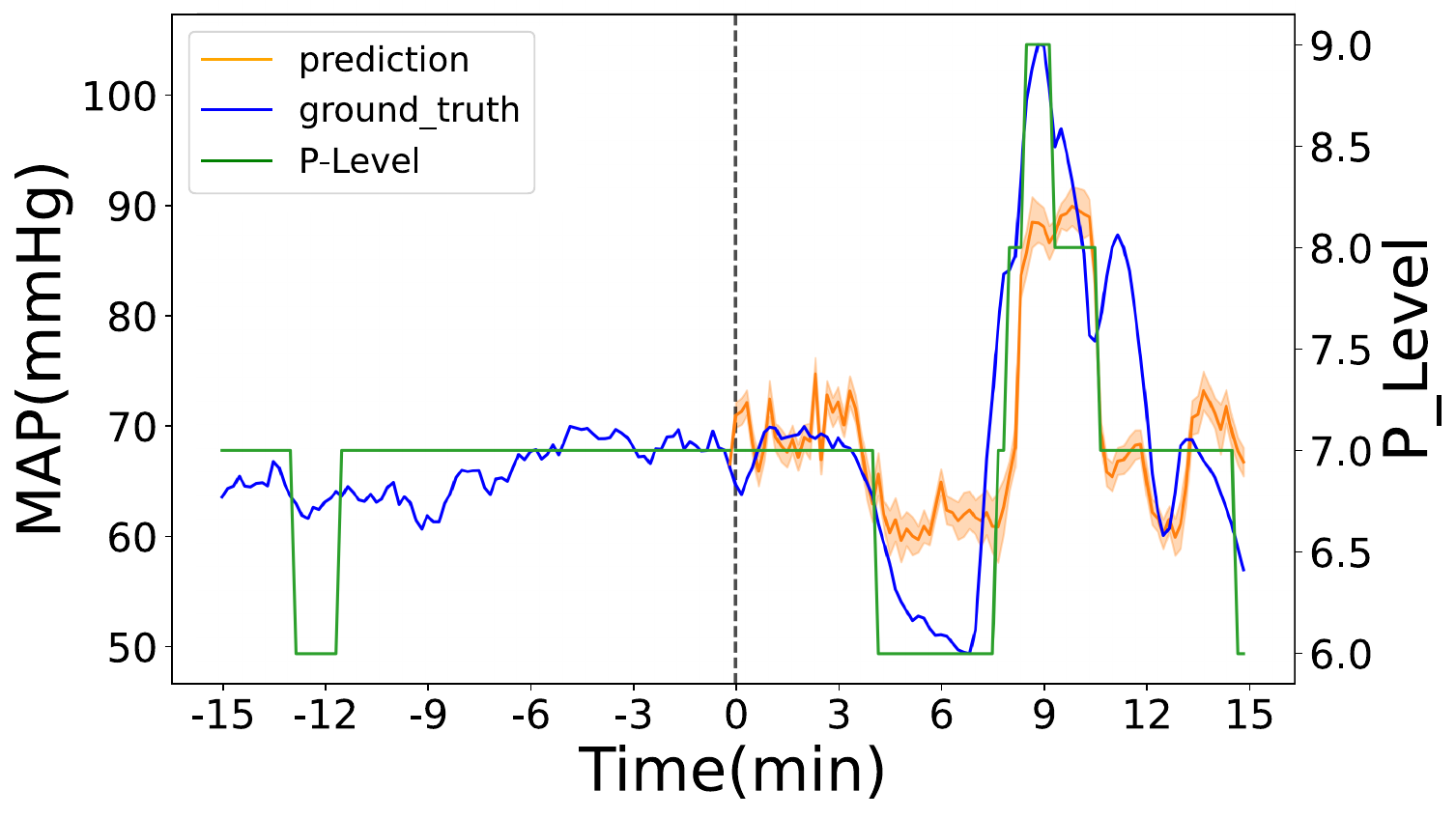}} \\

    \subfigure[NP \scriptsize{direct transfer}]{\includegraphics[width=0.32\linewidth]{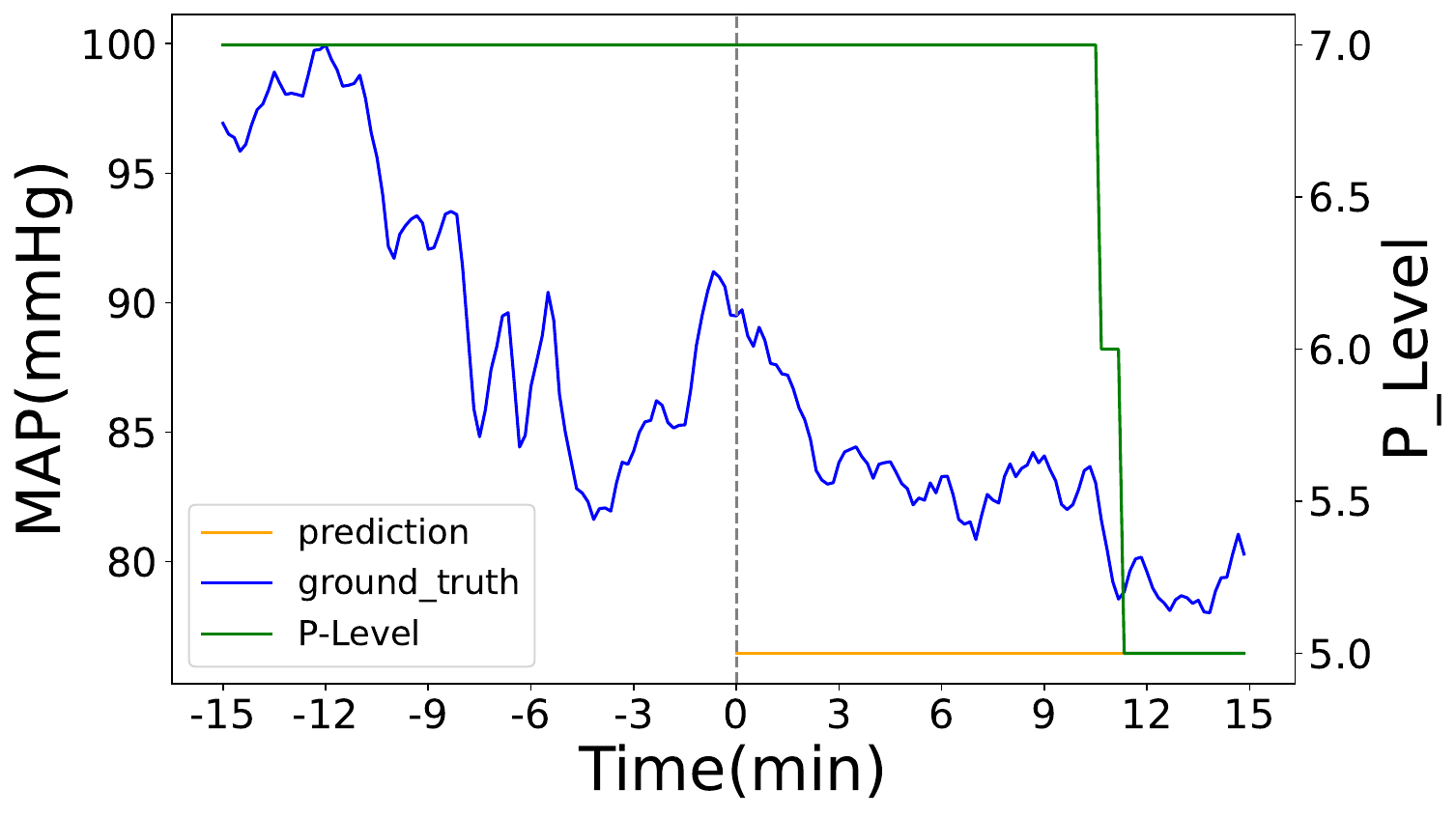}}
    \subfigure[CLMU]{\includegraphics[width=0.32\linewidth]{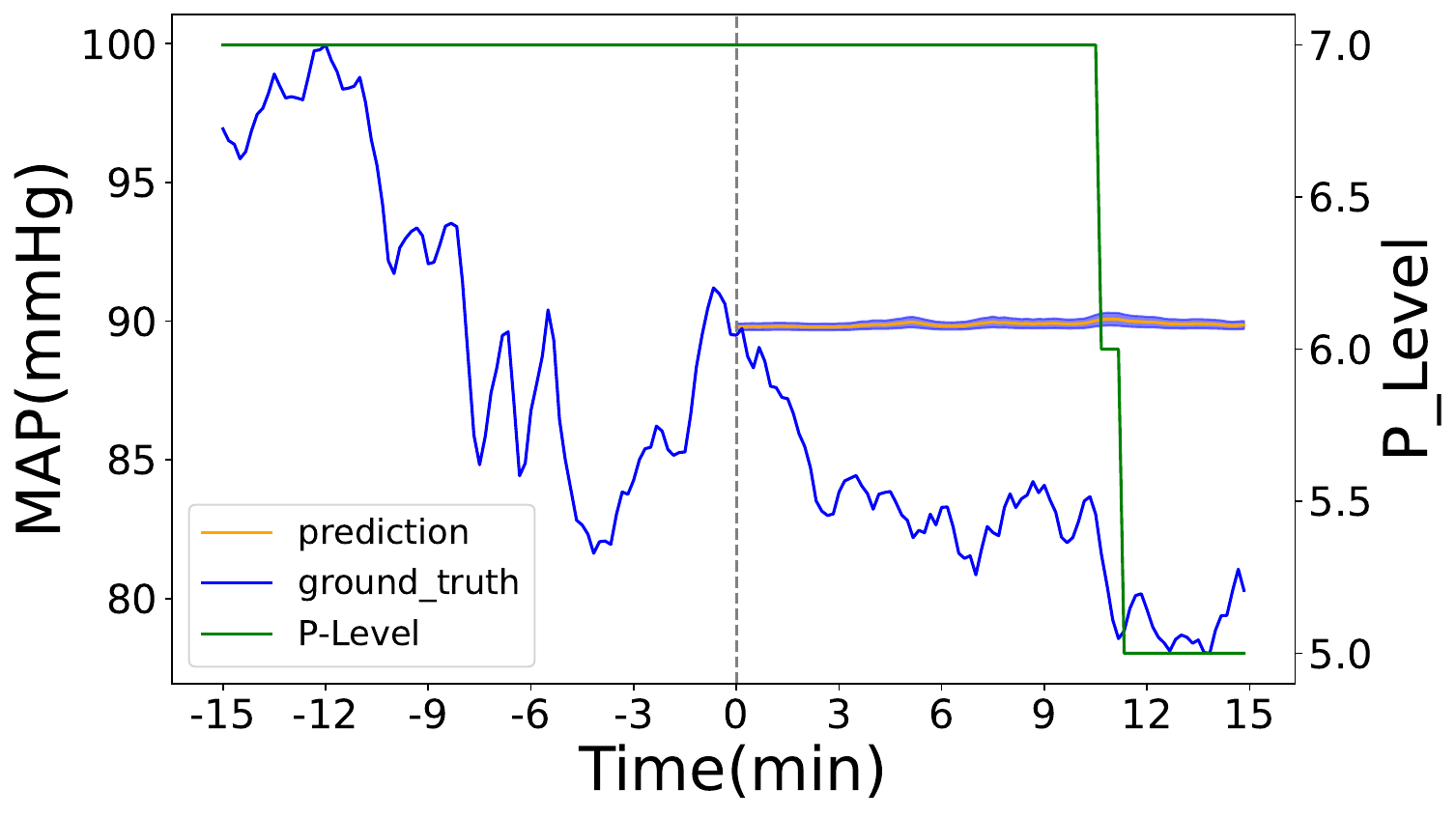}}
    \subfigure[DANP \textbf{(ours)}]{\includegraphics[width=0.32\linewidth]{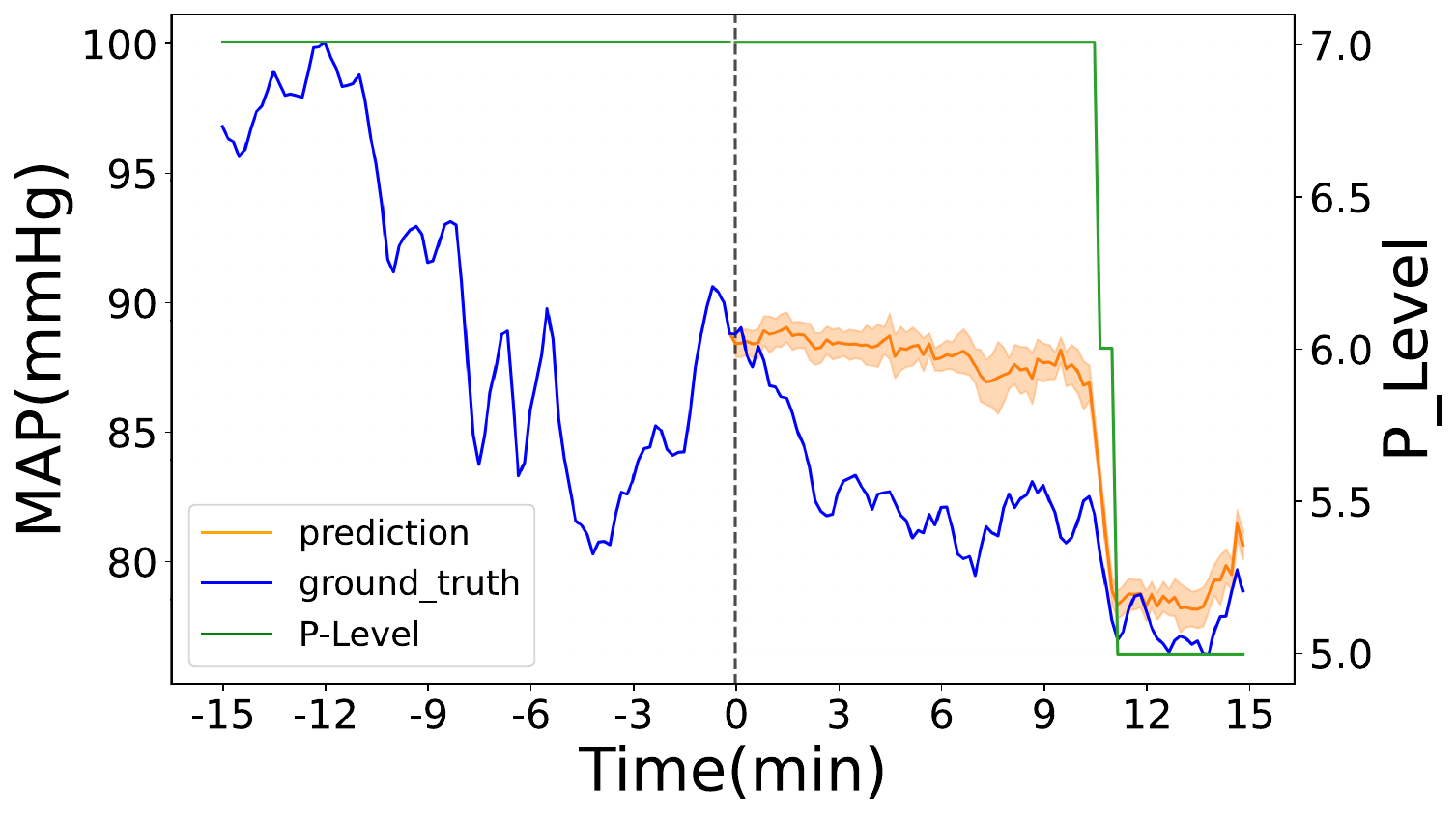}}
    
    \caption{Sample Prediction using Model Trained and Tested on HR-PCI Cohort. In scenarios where both direct transfer and CLMU can only produce flat forecasts, DANP is able to produce sensible MAP that corresponds to p-level shifts.}
    \label{fig:sample}
\end{figure}

 \paragraph{Scenario Creation.} We show two qualitative examples in Figure \ref{fig:sample} to illustrate DANP's ability for ``what-if" scenario creation. The first row (a)-(c) shows an increase of support level varying from P6-P9 and a simultaneous change in MAP. Only the DANP model is able to predict an increase in MAP in response to the increase in support. The second row (d)-(f) has a decreasing trend and support reduction from P7 to P5. Again, DT and CLMU predict a straight line and do not capture the trend over the forecast window, while the DANP captures the true trend in response to the p-level drop. These two samples illustrate the much-improved expressiveness of our DANP compared to existing baselines, and that DANP is able to produce sensible MAP in response to P-level shifts.

\subsection{Ablation Studies}

We compare DANP with other alternative approaches to address the distribution shift problem of learning from real datasets. For the neural process model (\texttt{NP no sim}), we experimented with the following modifications to address the challenges presented by our data.
\begin{itemize}
\itemsep0em 
\item \texttt{seq model:} The R-Encoder in our architecture encodes varying-length context into a representation vector, in addition to the neural-process-specific Z-Encoder.
\item \texttt{sampling:} subsampling training data to balance the trend distribution. \citep{branco2016survey} 
\item \texttt{DA:} domain adversarial training with simulation data, see section 4.1 for details.
\item \texttt{meta-regression:} We explore the meta-regression settings that are common to neural process baselines \citep{garnelo2018neural, nguyen2022transformer}, where each time series is considered a unique function with its own context and predictions, instead of learning the problem as one single function as in Equation \ref{eqn: problem} in our setting. In the meta-regression setting, $x$ of the neural process is the time stamp and P-level, and  $y$ is the hemodynamic features. The input time steps are used as context points to obtain the neural process that produces forecasts for future time steps. 
\end{itemize}

Our ablation experiment result shows that domain adversarial training (\texttt{DA}) allows the model to significantly improve in expressiveness, evident in the high trend accuracy and lower MAE values for increasing and decreasing trend data. Subsampling, an otherwise popular method to handle imbalanced data in machine learning, proved to severely harm prediction accuracy. This is potentially due to over-sampling noisy data in our datasets, which are prevalent among the non-stationary time series. The meta-regression model exhibits the behavior of capturing the dominant stationarity in the dataset and under-performing on the more clinically important non-stationary data.

 \begin{table}[]
     \centering
     \begin{tabular}{l c c c c c}
     \toprule
\textbf{Method} &  \textbf{MAE (mmHg)} $\downarrow$ &   \textbf{MAE (inc)} $\downarrow$ &  \textbf{MAE (dec)} $\downarrow$ & \textbf{MAE (stat)} $\downarrow$ &  \textbf{Trend Acc} $\uparrow$ \\ 
    \midrule
\texttt{NP \scriptsize{no sim}} & 8.68 \plusminus 0.06 & 6.90 \plusminus .01 & 15.34 \plusminus .02 &  7.63 \plusminus .01 & 0.52 \plusminus .00 \\
 \texttt{NP\scriptsize{ sim + real}} & 7.55 \plusminus .87 & 9.23 \plusminus .93 & 8.91 \plusminus .77 & 6.45 \plusminus .89 & 0.61 \plusminus .02 \\
\texttt{- seq model} & 7.19 \plusminus .10 & 8.73 \plusminus 1.26 & 10.28 \plusminus 1.42 & 5.92 \plusminus .02 & 0.04 \plusminus .05
\\
\texttt{+ sampling} &  8.84 \plusminus .72 & 9.49 \plusminus .91 &  11.86 \plusminus 2.05 & 7.85 \plusminus .64 & 0.45 \plusminus .18
\\
\texttt{+ DA + sampling} &  8.90 \plusminus .83 & 8.07 \plusminus .58 & 14.72 \plusminus 1.13 & 7.91 \plusminus .71 & 0.46 \plusminus .12
\\
\texttt{+ DA} (final) & 6.65 \plusminus .13 & \textbf{6.94} \plusminus .10 & \textbf{8.46} \plusminus .17 &   5.36 \plusminus .09 & \textbf{0.70} \plusminus .01
 \\
\texttt{meta-regression} & \textbf{5.57 }\plusminus 0.20 & 8.19
 \plusminus 1.23 & 9.74 \plusminus 1.06 & \textbf{4.23}
 \plusminus .95 & 0.70 \plusminus .08 \\
    \bottomrule
     \end{tabular}
     \caption{Ablation study on the current architecture.  The $+$ and $-$ signs indicate addition and deletion of the modifications. DANP (2nd to last row) was selected as our final model for its expressivity and prediction accuracy on non-stationary data.}
     \label{tab:ablation}
 \end{table}

\section{Conclusion and Discussion}
In this paper, we developed a data-driven simulator to predict MAP time series given different levels of pump speed in mechanical circulatory support (MCS) devices. Our innovation lies in the introduction of a Domain-Adversarial Neural Process (DANP), a generative probabilistic deep sequence model, which integrates both simulation data from the Lumped Parameter cardiac simulator with real-world sensor data from Impella CP.  This integrative approach not only facilitates simulating various scenarios, including those rarely observed in the data, but also allows the DANP model to be able to answer ``what-if'' questions regarding pump level changes and subsequent patient MAP response . This enhances DANP model's applicability in clinical settings for MCS patients.
Furthermore, incorporating both the diversity of simulation data and the expressiveness of the real-world data, the DANP model demonstrates superior accuracy, with a 5\% overall improvement and a 19\% increase in accuracy in non-stationary data compared to the best-performing baselines.  Its performance offers more insightful and expressive forecasts than existing models and simulators. These attributes make the DANP model an invaluable tool in scenarios where detecting dynamic and non-stationary change is crucial.

\paragraph{Future Works.}

Additional medical metrics are required to evaluate DANP's performance appropriately and assess its generalizability and robustness before applying them in clinical settings.

One direction to improve our model is to incorporate other modes of data to create a fuller picture of patient status to decide how much MCS support is needed. This could include doctor notes (text),  lab values (tabular style data),  or other important measures such as cardiac output, oxygen saturation, and lactate (measured less consistently at lower resolution). In our work, we rely on aortic pressure signals and heart rate as they are available as high-resolution signals.


Another interesting future direction is to explore the optimal decision-making problem using the DANP simulator to automatically recommend the MCS support levels. We plan to couple the data-driven simulator developed in this work with a reinforcement learning (RL) algorithm. The RL algorithm will interact with the simulator as the environment and optimize the policy of changing the MCS support levels for improved patient outcomes. 


\newpage
\acks{This work was supported in part  by a research grant from
Abiomed, Inc. (Danvers, MA), the U.S. Army Research Office
under Army-ECASE award W911NF-23-1-0231, the U.S. Department Of Energy, Office of Science, IARPA HAYSTAC Program, CDC-RFA-FT-23-0069, DARPA AIE FoundSci, NSF Grants \#2205093, \#2146343, and \#2134274.}
\bibliography{ref}

\begin{thebibliography}{30}
\providecommand{\natexlab}[1]{#1}
\providecommand{\url}[1]{\texttt{#1}}
\expandafter\ifx\csname urlstyle\endcsname\relax
  \providecommand{\doi}[1]{doi: #1}\else
  \providecommand{\doi}{doi: \begingroup \urlstyle{rm}\Url}\fi

\bibitem[AbioMed(2020)]{understanding}
AbioMed.
\newblock Automated impella controller™ (aic), 2020.
\newblock URL \url{https://www.heartrecovery.com/education/education-library/abiomed-automated-impella-controller-aic}.

\bibitem[Arndt et~al.(2020)Arndt, Hazara, Ghadirzadeh, and Kyrki]{arndt2020meta}
Karol Arndt, Murtaza Hazara, Ali Ghadirzadeh, and Ville Kyrki.
\newblock Meta reinforcement learning for sim-to-real domain adaptation.
\newblock In \emph{2020 IEEE International Conference on Robotics and Automation (ICRA)}, pages 2725--2731. IEEE, 2020.

\bibitem[B.~Geller(2022)]{escalation}
et.~al. B.~Geller.
\newblock Escalating and de-escalation temporary mechanical circulatory support in cardiogenic shock: A scientific statement from the american heart association.
\newblock \emph{Circulation}, 146:\penalty0 e50--68, August 2022.

\bibitem[Bonfanti et~al.(2017)Bonfanti, Balabani, Greenwood, Puppala, Homer-Vanniasinkam, and D{\'\i}az-Zuccarini]{bonfanti2017computational}
Mirko Bonfanti, Stavroula Balabani, John~P Greenwood, Sapna Puppala, Shervanthi Homer-Vanniasinkam, and Vanessa D{\'\i}az-Zuccarini.
\newblock Computational tools for clinical support: a multi-scale compliant model for haemodynamic simulations in an aortic dissection based on multi-modal imaging data.
\newblock \emph{Journal of The Royal Society Interface}, 14\penalty0 (136):\penalty0 20170632, 2017.

\bibitem[Branco et~al.(2016)Branco, Torgo, and Ribeiro]{branco2016survey}
Paula Branco, Lu{\'\i}s Torgo, and Rita~P Ribeiro.
\newblock A survey of predictive modeling on imbalanced domains.
\newblock \emph{ACM computing surveys (CSUR)}, 49\penalty0 (2):\penalty0 1--50, 2016.

\bibitem[Chen et~al.(2019)Chen, Chun, Patel, Chiang, and James]{chen2019validity}
Junqiao Chen, David Chun, Milesh Patel, Epson Chiang, and Jesse James.
\newblock The validity of synthetic clinical data: a validation study of a leading synthetic data generator (synthea) using clinical quality measures.
\newblock \emph{BMC medical informatics and decision making}, 19\penalty0 (1):\penalty0 1--9, 2019.

\bibitem[Cheng et~al.(2017)Cheng, Darnell, Dumitrascu, Chivers, Draugelis, Li, and Engelhardt]{cheng2017sparse}
Li-Fang Cheng, Gregory Darnell, Bianca Dumitrascu, Corey Chivers, Michael~E Draugelis, Kai Li, and Barbara~E Engelhardt.
\newblock Sparse multi-output gaussian processes for medical time series prediction.
\newblock \emph{arXiv preprint arXiv:1703.09112}, 2017.

\bibitem[Cheng et~al.(2019)Cheng, Orosz, Murray, and Burdick]{cheng2019end}
Richard Cheng, G{\'a}bor Orosz, Richard~M Murray, and Joel~W Burdick.
\newblock End-to-end safe reinforcement learning through barrier functions for safety-critical continuous control tasks.
\newblock In \emph{Proceedings of the AAAI conference on artificial intelligence}, volume~33, pages 3387--3395, 2019.

\bibitem[D.~Doshi(2015)]{harvi}
D.~Burkhoff D.~Doshi.
\newblock Cardiovascular simulation of heart failure pathophysiology and therapeutics.
\newblock \emph{Journal of Cardiac Failure}, 22:\penalty0 303--311, December 2015.

\bibitem[Davenport and Kalakota(2019)]{davenport2019potential}
Thomas Davenport and Ravi Kalakota.
\newblock The potential for artificial intelligence in healthcare.
\newblock \emph{Future healthcare journal}, 6\penalty0 (2):\penalty0 94, 2019.

\bibitem[Frangi et~al.(2018)Frangi, Tsaftaris, and Prince]{frangi2018simulation}
Alejandro~F Frangi, Sotirios~A Tsaftaris, and Jerry~L Prince.
\newblock Simulation and synthesis in medical imaging.
\newblock \emph{IEEE transactions on medical imaging}, 37\penalty0 (3):\penalty0 673--679, 2018.

\bibitem[Ganin et~al.(2016)Ganin, Ustinova, Ajakan, Germain, Larochelle, Laviolette, Marchand, and Lempitsky]{ganin2016domain}
Yaroslav Ganin, Evgeniya Ustinova, Hana Ajakan, Pascal Germain, Hugo Larochelle, Fran{\c{c}}ois Laviolette, Mario Marchand, and Victor Lempitsky.
\newblock Domain-adversarial training of neural networks.
\newblock \emph{The journal of machine learning research}, 17\penalty0 (1):\penalty0 2096--2030, 2016.

\bibitem[Garnelo et~al.(2018)Garnelo, Schwarz, Rosenbaum, Viola, Rezende, Eslami, and Teh]{garnelo2018neural}
Marta Garnelo, Jonathan Schwarz, Dan Rosenbaum, Fabio Viola, Danilo~J Rezende, SM~Eslami, and Yee~Whye Teh.
\newblock Neural processes.
\newblock \emph{arXiv preprint arXiv:1807.01622}, 2018.

\bibitem[Gottesman et~al.(2019)Gottesman, Johansson, Komorowski, Faisal, Sontag, Doshi-Velez, and Celi]{gottesman2019guidelines}
Omer Gottesman, Fredrik Johansson, Matthieu Komorowski, Aldo Faisal, David Sontag, Finale Doshi-Velez, and Leo~Anthony Celi.
\newblock Guidelines for reinforcement learning in healthcare.
\newblock \emph{Nature medicine}, 25\penalty0 (1):\penalty0 16--18, 2019.

\bibitem[Impella(2014)]{impella2014instructions}
CP~Impella.
\newblock Instructions for use \& clinical reference manual (united states only), abiomed.
\newblock \emph{Inc., Jul}, 2014.

\bibitem[Kingma and Welling(2013)]{kingma2013auto}
Diederik~P Kingma and Max Welling.
\newblock Auto-encoding variational bayes.
\newblock \emph{arXiv preprint arXiv:1312.6114}, 2013.

\bibitem[Li et~al.(2022)Li, Zhou, Jortberg, and Yu]{li2022forecasting}
Alan Li, Zihao Zhou, Elise Jortberg, and Rose Yu.
\newblock Forecasting aortic pressure cross-cohort with deep sequence models.
\newblock In \emph{2022 Computing in Cardiology (CinC)}, volume 498, pages 1--4. IEEE, 2022.

\bibitem[Mahmood et~al.(2018)Mahmood, Chen, Sudarsky, Yu, and Durr]{mahmood2018deep}
Faisal Mahmood, Richard Chen, Sandra Sudarsky, Daphne Yu, and Nicholas~J Durr.
\newblock Deep learning with cinematic rendering: fine-tuning deep neural networks using photorealistic medical images.
\newblock \emph{Physics in Medicine \& Biology}, 63\penalty0 (18):\penalty0 185012, 2018.

\bibitem[Morid et~al.(2023)Morid, Sheng, and Dunbar]{morid2023time}
Mohammad~Amin Morid, Olivia R~Liu Sheng, and Joseph Dunbar.
\newblock Time series prediction using deep learning methods in healthcare.
\newblock \emph{ACM Transactions on Management Information Systems}, 14\penalty0 (1):\penalty0 1--29, 2023.

\bibitem[Nguyen and Grover(2022)]{nguyen2022transformer}
Tung Nguyen and Aditya Grover.
\newblock Transformer neural processes: Uncertainty-aware meta learning via sequence modeling.
\newblock \emph{arXiv preprint arXiv:2207.04179}, 2022.

\bibitem[Prakosa et~al.(2012)Prakosa, Sermesant, Delingette, Marchesseau, Saloux, Allain, Villain, and Ayache]{prakosa2012generation}
Adityo Prakosa, Maxime Sermesant, Herv{\'e} Delingette, St{\'e}phanie Marchesseau, Eric Saloux, Pascal Allain, Nicolas Villain, and Nicholas Ayache.
\newblock Generation of synthetic but visually realistic time series of cardiac images combining a biophysical model and clinical images.
\newblock \emph{IEEE transactions on medical imaging}, 32\penalty0 (1):\penalty0 99--109, 2012.

\bibitem[Prasad et~al.(2017)Prasad, Cheng, Chivers, Draugelis, and Engelhardt]{prasad2017reinforcement}
Niranjani Prasad, Li-Fang Cheng, Corey Chivers, Michael Draugelis, and Barbara~E Engelhardt.
\newblock A reinforcement learning approach to weaning of mechanical ventilation in intensive care units.
\newblock \emph{arXiv preprint arXiv:1704.06300}, 2017.

\bibitem[Roberts et~al.(2013)Roberts, Osborne, Ebden, Reece, Gibson, and Aigrain]{roberts2013gaussian}
Stephen Roberts, Michael Osborne, Mark Ebden, Steven Reece, Neale Gibson, and Suzanne Aigrain.
\newblock Gaussian processes for time-series modelling.
\newblock \emph{Philosophical Transactions of the Royal Society A: Mathematical, Physical and Engineering Sciences}, 371\penalty0 (1984):\penalty0 20110550, 2013.

\bibitem[Schampaert et~al.(2011)Schampaert, van't Veer, van~de Vosse, Pijls, de~Mol, and Rutten]{schampaert2011vitro}
St{\'e}phanie Schampaert, Marcel van't Veer, Frans~N van~de Vosse, Nico~HJ Pijls, Bas~A de~Mol, and Marcel~CM Rutten.
\newblock In vitro comparison of support capabilities of intra-aortic balloon pump and impella 2.5 left percutaneous.
\newblock \emph{Artificial Organs}, 35\penalty0 (9):\penalty0 893--901, 2011.

\bibitem[Schulam and Saria(2015)]{schulam2015framework}
Peter Schulam and Suchi Saria.
\newblock A framework for individualizing predictions of disease trajectories by exploiting multi-resolution structure.
\newblock \emph{Advances in neural information processing systems}, 28, 2015.

\bibitem[T.~Heldt(2001)]{windkessel}
et.~al. T.~Heldt.
\newblock Computational modeling of cardiovascular response to orthostatic stress.
\newblock \emph{J Appl Physiol}, 92:\penalty0 1239--1254, 2001.

\bibitem[Traor{\'e} et~al.(2019)Traor{\'e}, Caselles-Dupr{\'e}, Lesort, Sun, D{\'\i}az-Rodr{\'\i}guez, and Filliat]{traore2019continual}
Ren{\'e} Traor{\'e}, Hugo Caselles-Dupr{\'e}, Timoth{\'e}e Lesort, Te~Sun, Natalia D{\'\i}az-Rodr{\'\i}guez, and David Filliat.
\newblock Continual reinforcement learning deployed in real-life using policy distillation and sim2real transfer.
\newblock \emph{arXiv preprint arXiv:1906.04452}, 2019.

\bibitem[van~de Meerakker et~al.(2019)van~de Meerakker, van~de Vosse, Rutten, van Dort, Foolen, and Peij]{van2019comparing}
T~van~de Meerakker, F~van~de Vosse, M~Rutten, D~van Dort, J~Foolen, and K~Peij.
\newblock Comparing heart assist devices, using a lumped parameter model and mock-loop.
\newblock 2019.

\bibitem[Voelker et~al.(2019)Voelker, Kaji{\'c}, and Eliasmith]{voelker2019legendre}
Aaron Voelker, Ivana Kaji{\'c}, and Chris Eliasmith.
\newblock Legendre memory units: Continuous-time representation in recurrent neural networks.
\newblock \emph{Advances in neural information processing systems}, 32, 2019.

\bibitem[Zhao et~al.(2020)Zhao, Queralta, and Westerlund]{zhao2020sim}
Wenshuai Zhao, Jorge~Pe{\~n}a Queralta, and Tomi Westerlund.
\newblock Sim-to-real transfer in deep reinforcement learning for robotics: a survey.
\newblock In \emph{2020 IEEE symposium series on computational intelligence (SSCI)}, pages 737--744. IEEE, 2020.

\end{thebibliography}

\newpage
\appendix
\section{Data}
\label{app:data}

\textbf{Deterministic Cardiovascular Simulator ranges}: P-levels were varied across all settings of the Impella CP (P1-P9), heart rate values were varied from 40 - 300 bpm, end-systolic elastance was varied from 0.2 - 10 mmHg/mL, and time constant of relaxation values were varied from 0.015 - 0.12 sec.

\begin{figure*}[hb]
\centering
\includegraphics[width=0.7\textwidth]{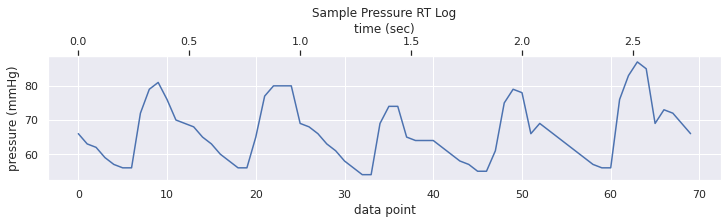}
\caption{Sample Pump Performance (25 Hz) Data }
\label{RT25}
\end{figure*}

\paragraph{Categorizing trends in MAP data.} We conduct experiments to predict MAP as a function of pump speed and other physiological features. In order to assess the model's ability to handle non-stationary data, we categorize data into decrease (dec), increase (inc), and stationary (stat) subsets.  The trend of input data is determined by fitting a simple linear regression line. The midpoint value of this regression line, termed `$\text{medium}$', and the overall rate of change, `$dy$', are calculated across input data. If `$\text{medium}$' exceeds 80mmHg, trends are classified as ``inc'' for a `$dy$' greater than or equal to 10mmHg, ``dec'' for a `$dy$' less than or equal to -10mmHg and ``stat'' otherwise. For a `$\text{medium}$' of 80 or below, the thresholds for ``inc'' and ``dec'' are adjusted to 5 and -5 mmHg, with any other rate being classified as ``stat''. These were chosen as approximations for concerning rates of change in a given MAP range. 

\section{Additional Experimental results}
\label{app:exp}

\subsection{Learned data distribution}
\begin{figure*}[h]
    \centering
    \subfigure[LP Model]{\includegraphics[height=0.25\linewidth]{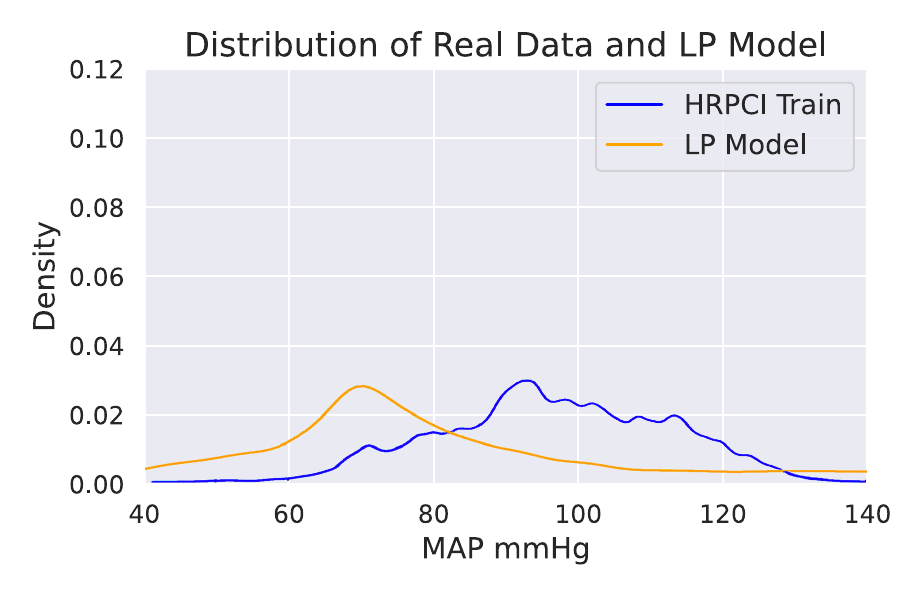} }
    \subfigure[DANP Model]{\includegraphics[height=0.25\linewidth]{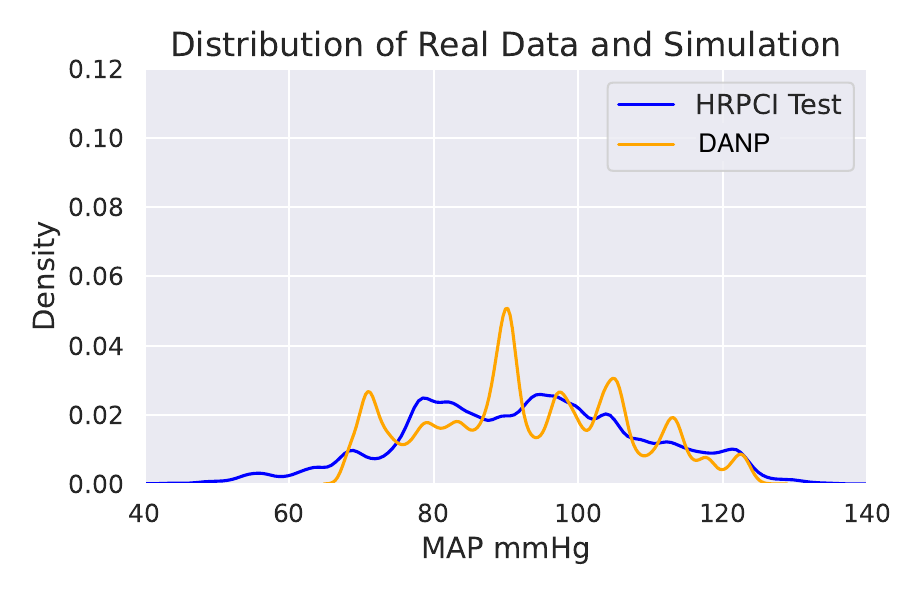}}
    \caption{Distributions of Cohort Data vs Simulated Data from LP Model and our proposed DANP model}
        \label{fig:dist}%
\end{figure*}

\begin{figure*}[h]
    \centering
    \subfigure[Direct Transfer]{\includegraphics[height=0.25\linewidth]{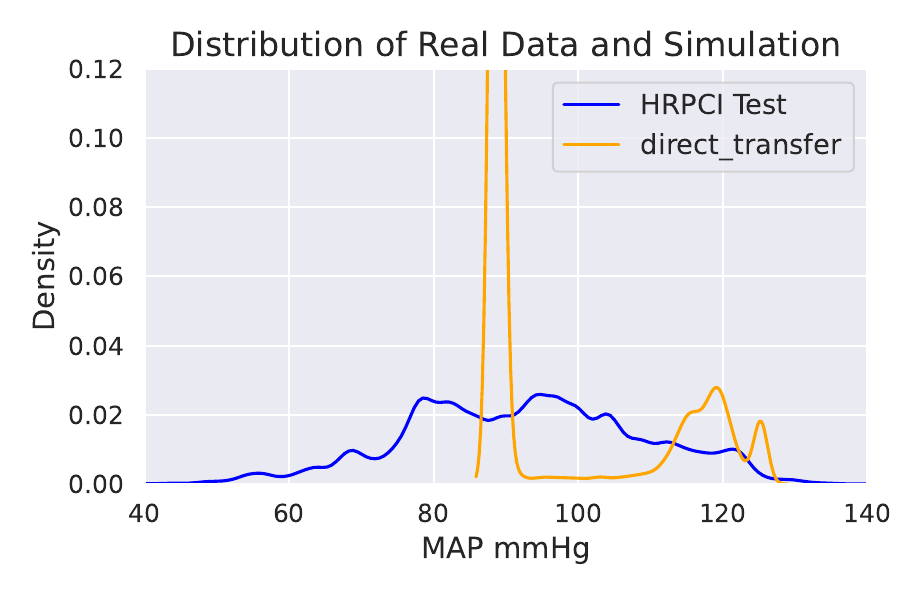} }
    \subfigure[CLMU Model]{\includegraphics[height=0.25\linewidth]{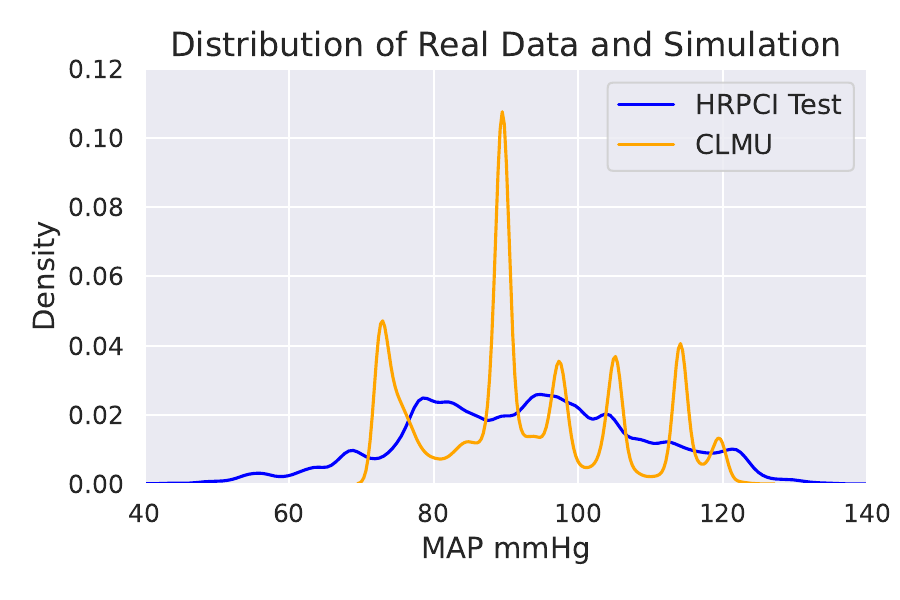} }
    \caption{Distributions of Cohort Data from Direct Transfer and CLMU}
        \label{fig:dist2}%
\end{figure*}

We compared the distribution of MAP values simulated by the LP model with that learned by the LP Model, Direct Transfer, CLMU, and DANP across four visuals: (1)
\emph{LP Model (Figure \ref{fig:dist} panel a):} Real train data peaks at 90-100 mmHg, while the LP Model's simulated data peaks at 70-90 mmHg.
(2) \emph{Direct Transfer (Figure \ref{fig:dist2} panel a):} It shows multiple peaks with minimal overlap with ground truth test data, suggesting lower accuracy.
(3) \emph{CLMU Model (Figure \ref{fig:dist2} panel b):} Multiple peaks are observed, with the dominant peak around 90 mmHg. The overlap with the ground truth test data highlights its improved representation.
(4) \emph{DANP (Figure \ref{fig:dist} panel b):} DANP provides the most accurate results, with a smooth distribution. Its main peak at 90 mmHg mirrors real train data, showcasing its ability to reflect real data precisely and its superior overlap with actual data.

\end{document}